\documentclass{article}

\usepackage{arxiv}

\usepackage[utf8]{inputenc}
\usepackage[T1]{fontenc}

\usepackage{amsmath}
\usepackage{amsfonts}
\usepackage{amssymb}
\usepackage{natbib}
\usepackage{doi}
\usepackage{url}
\usepackage{array}
\usepackage{booktabs}
\usepackage{multirow}

\usepackage{graphicx}
\usepackage{xcolor}

\usepackage{nicefrac}
\usepackage{microtype}

\usepackage{natbib}
\usepackage{doi}
\usepackage{url}
\usepackage{hyperref}

\renewcommand{\shorttitle}{One Step Is Optimal}

\title{One Step Is Optimal: Unconditional Rectified Flows as Noise2Noise Denoisers for Low-Dose CT}

\author{
{Timothy Sereda} \\
Department of Computer Science\\
Biomedical Perception \& Intelligence Lab\\
University of South Dakota\\
Vermillion, South Dakota, USA\\
\texttt{timothy.sereda@coyotes.usd.edu}
\And
\href{https://orcid.org/0000-0002-8078-6730}
{\includegraphics[scale=0.06]{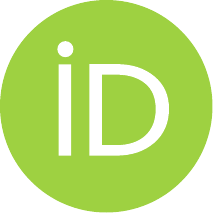}\hspace{1mm}Debesh Jha} \\
Department of Computer Science\\
Biomedical Perception \& Intelligence Lab\\
University of South Dakota\\
Vermillion, South Dakota, USA\\
\texttt{debesh.jha@usd.edu}
}

\begin{document}
\maketitle

\begin{abstract}
We study rectified flows for denoising without clean training targets.
For an unconditional rectified flow trained on Noise2Noise pairs, the optimal velocity field yields the minimum mean squared error (MMSE) denoiser in one step. In a jointly Gaussian model, we show that integrating the flow before applying its denoising readout increases error with integration time. The flow
transports between noisy distributions, so its endpoint is not a clean-image
estimate. Experiments on low-dose CT and Gaussian-corrupted natural images
follow this pattern: one-step flow performs similarly to direct regression on
the same pairs, whereas additional Euler steps reduce PSNR. A direct regressor
using spatially separated, similar patches gives our best label-free CT result,
a $1.99$\,dB gain over the noisy input. We report these results in CTDenoiser, a
benchmark covering five architectures, supervised and self-supervised training,
and two per-image methods. Supervised CNNs gain about $4$\,dB, while label-free
performance depends strongly on how training pairs account for correlated CT
noise. A channelized Hotelling observer also shows that fidelity gains need not
improve lesion detectability: detectability falls at $160$\,HU despite higher
PSNR and SSIM, while changes at $40$--$80$\,HU are small. Code, model definitions, and data-loading utilities will be released publicly; the repository link is
withheld for anonymous review.
\end{abstract}

\section{Introduction}

Flow-based denoisers learn a sequence of transformations from a noisy image to a
target distribution. When the targets are also noisy, it is unclear whether
following this sequence improves on a single regression step. We study this
question for unconditional rectified flows trained on Noise2Noise pairs.

At the population optimum, a single evaluation of the velocity field recovers
the posterior mean of the clean signal. In a jointly Gaussian model, integrating
the flow before applying the denoising readout moves the estimate away from this
minimum mean squared error (MMSE) solution and toward the noisy input. Our
experiments show a corresponding decline in PSNR as the number of Euler steps
increases. On the same noisy pairs, direct regression performs slightly better
than one-step flow and requires no integration.

Low-dose computed tomography (LDCT) is a useful setting for this comparison.
Reducing radiation exposure limits the risks associated with CT
imaging~\cite{brenner2007computed}, but increases image noise. Denoising must
suppress that noise while preserving anatomical detail. Convolutional
encoder-decoders~\cite{chen2017lowdose}, transformers~\cite{wang2022ctformer},
and generative models have all been applied to this problem. Training without
paired full-dose targets remains difficult because CT reconstruction produces
spatially correlated noise, violating the pixel-independence assumption used by
many blind-spot and downsampling methods.

Comparisons are further complicated by differences in preprocessing,
normalisation, evaluation regions, and metric
implementations~\cite{eulig2024benchmark}. We use CTDenoiser to evaluate five
architectures with a shared training and evaluation pipeline on the TCIA LDCT
dataset~\cite{moen2021low}. The benchmark covers supervised, similarity-based,
and blind-spot training, together with two per-image methods. It uses
patient-level splits and full-slice evaluation with overlapping tiles. We
report PSNR, SSIM, RMSE, gradient magnitude similarity deviation (GMSD), and a
residual spectral ratio (RSR), alongside a task-based detectability study.

We derive the one-step Noise2Noise
identity for an unconditional rectified flow and prove increasing denoising
risk along its readout in a jointly Gaussian model
(Section~\ref{sec:method}). We compare flow and regression using the same
similarity-based pairs, and test the effect of additional inference steps on CT
and natural images. Finally, we assess lesion detectability with a channelized
Hotelling observer (Section~\ref{sec:detect}). The observer shows that higher
PSNR and SSIM can accompany reduced detectability, with the largest loss at the
highest lesion contrast tested.

% -----------------------------------------------------------------------

\section{Related Work}

\paragraph{CNN and transformer denoisers.}
RED-CNN~\cite{chen2017lowdose} uses a residual encoder-decoder with multi-level
skip connections. DnCNN~\cite{zhang2017beyond} predicts noise with a stack of
Conv-BN-ReLU layers and subtracts it from the input. CTformer~\cite{wang2022ctformer}
uses a U-shaped transformer with Token2Token Dilation (T2TD) blocks, combining
dilated convolutions, cyclic spatial shifts, and attention to aggregate local
and longer-range information.

\paragraph{Generative approaches.}
Score-based diffusion models~\cite{song2021scorebased} have been applied to CT
denoising~\cite{gao2023corediff}, although iterative sampling can make inference
expensive. Flow matching~\cite{lipman2022flow,albergo2022building} learns a
velocity field between source and target distributions. Poisson Flow
Consistency Models (PFCM)~\cite{hein2025pfcm} distil a generative LDCT denoiser
to a single step and examine variability in the generated images.

These methods raise a distinction between perceptual quality and fidelity to
the measured image, formalised by the perception--distortion
trade-off~\cite{blau2018perception}. The connection between flow matching and
denoising is also established: the optimal velocity can be used to obtain a
conditional-mean estimate~\cite{gagneux2025phases,albergo2023stochastic}.
Posterior-mean rectified flow (PMRF)~\cite{ohayon2025pmrf} and inversion by
direct iteration (InDI)~\cite{delbracio2023indi} relate restoration to
generative transport. We examine the conditional-mean identity when both
endpoints are noisy observations and compare the resulting one-step estimator
with multi-step inference.

\paragraph{Self-supervised and zero-shot denoisers.}
Noise2Noise~\cite{lehtinen2018noise2noise} shows that regression between
independent, conditionally zero-mean noisy observations recovers the same
conditional mean as regression to clean targets. Noise2Void~\cite{krull2019noise2void}
uses a blind spot to withhold the target pixel from the network, requiring only
one noisy image but assuming pixel-wise independent noise.
Neighbor2Neighbor~\cite{huang2021neighbor2neighbor} constructs pairs by
subsampling neighbouring pixels. Spatially correlated CT noise makes these
independence assumptions difficult to satisfy.

Noise2Sim~\cite{niu2020noise2sim} instead pairs the centre pixels of non-local,
similar patches. Asymmetric pixel-shuffle downsampling
(AP-BSN)~\cite{lee2022apbsn} reduces spatial noise correlation through a larger
training stride. Other methods train on individual images: Zero-Shot
Noise2Noise (ZS-N2N)~\cite{mansour2023zsn2n} fits a small network to a
downsampled pair, while Filter2Noise~\cite{sun2025filter2noise} uses an
attention-guided bilateral filter for LDCT. WIA-LD2ND~\cite{zhao2024wia}
performs self-supervised LDCT denoising in the wavelet domain.
Noise2Score~\cite{kim2021noise2score} approaches the problem through score
estimation and Tweedie's formula for the posterior mean. Our benchmark includes
Noise2Void, Noise2Sim, ZS-N2N, and Filter2Noise.

\paragraph{Task-based and hallucination-aware evaluation.}
PSNR and SSIM can disagree with assessments of restoration quality on medical
images~\cite{breger2025reassess} and with radiologist assessments of denoised
CT~\cite{lee2025ldctiqa}. Model observers, including the channelized Hotelling
observer (CHO), provide task-based measures such as detectability ($d'$) and
area under the ROC curve. Li et al.~\cite{li2021detection} showed that denoising
can remove information needed for a detection task even when conventional
image-quality metrics improve. Kc and Zeng~\cite{kc2024lgcho} reported a
similar discrepancy between fidelity metrics and low-contrast detectability
in CT.
\label{sec:rw-taskbased}

Hallucination measures address a related concern. Bhadra et
al.~\cite{bhadra2021halluc} defined hallucinations through a measurement and
null-space decomposition; Tivnan et al.~\cite{tivnan2024halluc} proposed a
Hallucination Index for generative reconstruction. These studies motivate our
use of signal-present and signal-absent observer tests in
Section~\ref{sec:detect}.

\paragraph{Evaluation metrics.}
PSNR and SSIM measure fidelity to a reference. Perceptual metrics such as
LPIPS~\cite{zhang2018unreasonable} assess visual similarity, while the noise
power spectrum (NPS)~\cite{solomon2012characterization} describes noise texture.
These measures address different aspects of denoising quality and complement
task-based evaluation.

% -----------------------------------------------------------------------

\section{Methods}

\subsection{Data and Preprocessing}

We use the TCIA LDCT-and-Projection-Data
dataset~\cite{moen2021low}, which contains paired low-dose and full-dose
reconstructions.  DICOM slices are read with \texttt{pydicom}, sorted by
$z$-position, and converted to Hounsfield units (HU).  HU values are
windowed to $[0, 1]$ via

\begin{equation}
  x = \operatorname{clamp}\!\left(\frac{u + b}{s},\; 0,\; 1\right),
  \label{eq:hu-norm}
\end{equation}

where $u$ is the raw HU value and $(b,s)$ is a fixed anatomy window preset.
The processed arrays are cached as an HDF5 file (one dataset per
patient/dose level) for efficient random access during training.

Patients are assigned to training and validation sets using a fixed random
seed and a configurable validation fraction
($\text{val\_fraction}=0.2$ by default). All slices from a patient remain in
the same set.  During
training, $64\times64$ patches are sampled uniformly at random from each
slice.  During validation, full CT slices of variable size are processed
via the overlapped-tiling inference strategy described in
Section~\ref{sec:inference}.

\subsection{Architectures}

\paragraph{RED-CNN.}
Chen et al.'s~\cite{chen2017lowdose} Residual Encoder-Decoder CNN uses
$5\times5$ convolutions throughout.  The encoder consists of five
convolutional layers (96 filters each), and the decoder mirrors this
with five transposed-convolutional layers.  Skip connections are added at
three depths, and all activations use ReLU.

\paragraph{DnCNN.}
Zhang et al.'s~\cite{zhang2017beyond} denoising CNN uses 17 layers of
$3\times3$ convolutions.  The first layer has no batch normalisation;
intermediate layers each apply Conv-BN-ReLU; the final layer has neither.
The network predicts the noise component $n$ and returns
$\hat{x} = x - n$, a residual-learning formulation.

\paragraph{CTformer.}
Wang et al.'s~\cite{wang2022ctformer} architecture tokenises the input
with a strided convolution, applies a symmetric
encoder-bottleneck-decoder of Transformer blocks interleaved with
Token2Token Dilation (T2TD) modules, then detokenises with a transposed
convolution.  Each T2TD block applies a cyclic spatial shift and a dilated
re-tokenisation before multi-head self-attention.

\paragraph{U-Net.}
Our U-Net baseline uses three encoder stages, a two-layer bottleneck, and a
symmetric decoder with bilinear upsampling and concatenated skip
connections.  It shares the overall structure of the CFM velocity network, allowing a
comparison of MSE regression and flow matching with similar backbones.

\paragraph{Conditional Flow Matching (CFM).}
\label{sec:cfm}
Conditional flow matching~\cite{lipman2022flow} frames denoising as
learning a time-dependent vector field $v_\theta$ that transports a
source distribution $p_0$ (LDCT images) to a target distribution $p_1$
(full-dose images).  The \emph{rectified flow}
formulation~\cite{liu2022flow} uses straight-line paths, so the
conditional vector field is simply

\begin{equation}
  v^* = x_1 - x_0,
  \label{eq:rectified-target}
\end{equation}

constant along the path.  The training objective is

\begin{equation}
  \mathcal{L}_\text{CFM}(\theta)
  = \mathbb{E}_{t, x_0, x_1}\bigl[\|v_\theta(x_t,\, x_0,\, t) - (x_1 - x_0)\|^2\bigr],
  \label{eq:cfm-loss}
\end{equation}

where $t \sim \mathcal{U}[0,1]$, the interpolant is
$x_t = (1-t)x_0 + t\,x_1$, and $x_0$ (the noisy LDCT patch) is passed as a
conditioning channel: the velocity network input is the two-channel
concatenation $[x_t; x_0]$.  At inference we integrate the learned ODE with
$K$ forward-Euler steps initialised at the noisy input.  We train this conditional formulation on paired full-dose targets and use it
as the supervised flow baseline for Section~\ref{sec:method}.

\subsection{Training}

All models use the Adam optimiser at learning rate $10^{-4}$ and batch
size 4 (patches of $64\times64$).  Deterministic models (RED-CNN, DnCNN,
CTformer, U-Net) minimise pixel-level MSE; CFM uses
$\mathcal{L}_\text{CFM}$ (Eq.~\eqref{eq:cfm-loss}).

\paragraph{Supervision regimes.}
\emph{Supervised} training uses paired full-dose targets.
\emph{Noise2Void}~\cite{krull2019noise2void} replaces a fraction
$\rho=0.02$ of pixels per patch with a random neighbour within radius
$r=2$ and computes the loss against the original noisy value at those
blind-spot locations only.  \emph{Noise2Sim}~\cite{niu2020noise2sim}
constructs a self-supervised target from the centre pixels of non-local
self-similar patches.  \emph{ZS-N2N}~\cite{mansour2023zsn2n} is
architecture-agnostic: for each image a fresh three-layer residual network
($21.3\mathrm{K}$ parameters) is optimised for 2000 iterations on a
diagonally-downsampled Noise2Noise pair, then discarded.
\emph{Filter2Noise}~\cite{sun2025filter2noise} similarly optimises a small
single-image filter at test time.  The CFM baseline is evaluated only with paired full-dose targets; we do not
include blind-spot or similarity-based variants of this conditional model.

\subsection{Inference}
\label{sec:inference}

Full CT slices are generally larger than the training patches. We process each
slice with overlapping $P\times P$ tiles, using a margin $m=P/4$ and a weighted
mean to blend predictions in the overlap. Margins are discarded only where
tiles share an edge; tiles at the image boundary retain their full extent.
An earlier implementation also discarded the outer image margins, leaving a
zero-valued border that biased model metrics relative to the untiled identity
baseline. All results reported here use the corrected boundary handling.

\subsection{Evaluation Metrics}

\paragraph{PSNR / SSIM / RMSE.}
PSNR uses data range $R=1$ after normalisation; SSIM~\cite{wang2004image}
uses a Gaussian window ($\sigma=1.5$, window $11$); RMSE is
$\sqrt{\text{MSE}}$.

\paragraph{Gradient Magnitude Similarity Deviation (GMSD).}
GMSD~\cite{xue2014gradient} computes gradient magnitudes of the predicted
and reference images with Prewitt kernels and reports the standard
deviation of the gradient-magnitude-similarity map; lower is better and no
pretrained features are required.

\paragraph{Residual Spectral Ratio (RSR).}
\begin{equation}
  \text{RSR} = \frac{\mathbb{E}\bigl[|\mathcal{F}(\hat{x}-x_1)|^2\bigr]}
                    {\mathbb{E}\bigl[|\mathcal{F}(x_1)|^2\bigr] + \varepsilon},
  \label{eq:nps}
\end{equation}
where $\mathcal{F}$ is the 2-D DFT. RSR measures residual energy relative to
reference-image energy in the Fourier domain. It is an aggregate error ratio,
rather than a description of the spatial frequencies present in the residual.
The conventional CT noise-power spectrum (NPS) characterises noise estimated
from uniform-region ROIs. We report its radial centroid separately in
Section~\ref{sec:detect}.

% -----------------------------------------------------------------------

\section{Experiments}

\subsection{Setup}

Experiments use the TCIA LDCT-and-Projection-Data dataset~\cite{moen2021low}
(abdomen window) with patient-level splits ($\sim$$6{,}500$ training and
$\sim$$1{,}600$ validation slices per split).  The main benchmark (Table~\ref{tab:results}) reports the mean over three such
splits. Split and seed counts for the ablations and per-image experiments are
given in their table captions.  Trained models
run for 50 epochs with patch size 64, batch size 4, and Adam at $10^{-4}$; the
CFM model integrates the ODE with $K=20$ Euler steps during training and 5 at
evaluation.  The per-image methods (ZS-N2N, Filter2Noise) optimise one
network per validation image and are model-agnostic.  Validation metrics
are computed over full slices using the corrected overlapped inference of
Section~\ref{sec:inference} with margin $P/4 = 16$.  We report metric gains ($\Delta$) relative to the unprocessed LDCT input.

\subsection{Results}
\label{sec:results}

\begin{table*}[t]
  \centering
  \small
  \caption{TCIA LDCT abdomen benchmark: full-slice validation over three patient-level
splits after 50 epochs, using corrected tiling. PSNR is shown with its gain
over each split's noisy input (mean $30.25$\,dB); uncertainties are standard
deviations across splits. CFM is evaluated only under supervision. Per-image
methods and SSFlow ablations appear in Tables~\ref{tab:sweep-periimage}
and~\ref{tab:ssflow-steps}. Configuration: \texttt{sweep.yml}.
\textbf{Bold}: best $\Delta$PSNR within each supervision regime.
$\uparrow$/$\downarrow$: higher/lower is better.}
  \label{tab:results}
  \begin{tabular}{llrccccc}
    \toprule
    \textbf{Model} & \textbf{Supervision} & \textbf{Params} &
    PSNR\,$\uparrow$ (\,$\Delta$) & SSIM\,$\uparrow$ & RMSE\,$\downarrow$ &
    GMSD\,$\downarrow$ & RSR\,$\downarrow$ \\
    \midrule
    LDCT input & --- (no denoising) & --- & $30.25$ & $0.876$ & $0.0320$ & $0.174$ & $0.0115$ \\
    \midrule
    RED-CNN~\cite{chen2017lowdose}   & Supervised & $1.85\mathrm{M}$ & \textbf{34.50 \,($+4.25{\pm}0.23$)} & $0.927{\pm}0.013$ & $0.0194$ & $0.167$ & $0.0041$ \\
    U-Net                            & Supervised & $1.95\mathrm{M}$ & 34.26 \,($+4.02{\pm}0.22$) & $0.925{\pm}0.014$ & $0.0199$ & $0.188$ & $0.0044$ \\
    DnCNN~\cite{zhang2017beyond}     & Supervised & $0.56\mathrm{M}$ & 34.23 \,($+3.99{\pm}0.18$) & $0.924{\pm}0.014$ & $0.0200$ & $0.198$ & $0.0044$ \\
    CFM~\cite{lipman2022flow}        & Supervised & $2.11\mathrm{M}$ & 33.09 \,($+2.84{\pm}0.22$) & $0.915{\pm}0.013$ & $0.0228$ & $0.192$ & $0.0057$ \\
    CTformer~\cite{wang2022ctformer} & Supervised & $0.40\mathrm{M}$ & 32.01 \,($+1.77{\pm}0.35$) & $0.904{\pm}0.013$ & $0.0255$ & $0.249$ & $0.0070$ \\
    \midrule
    U-Net                            & Noise2Sim~\cite{niu2020noise2sim} & $1.95\mathrm{M}$ & \textbf{32.08 \,($+1.83{\pm}0.08$)} & $0.901{\pm}0.015$ & $0.0259$ & $0.174$ & $0.0075$ \\
    RED-CNN~\cite{chen2017lowdose}   & Noise2Sim~\cite{niu2020noise2sim} & $1.85\mathrm{M}$ & 32.05 \,($+1.81{\pm}0.10$) & $0.901{\pm}0.015$ & $0.0259$ & $0.163$ & $0.0075$ \\
    DnCNN~\cite{zhang2017beyond}     & Noise2Sim~\cite{niu2020noise2sim} & $0.56\mathrm{M}$ & 31.93 \,($+1.69{\pm}0.09$) & $0.898{\pm}0.015$ & $0.0263$ & $0.187$ & $0.0078$ \\
    CTformer~\cite{wang2022ctformer} & Noise2Sim~\cite{niu2020noise2sim} & $0.40\mathrm{M}$ & 31.04 \,($+0.79{\pm}0.13$) & $0.887{\pm}0.018$ & $0.0289$ & $0.232$ & $0.0092$ \\
    \midrule
    RED-CNN~\cite{chen2017lowdose}   & Noise2Void~\cite{krull2019noise2void} & $1.85\mathrm{M}$ & \textbf{30.67 \,($+0.42{\pm}0.06$)} & $0.883{\pm}0.018$ & $0.0303$ & $0.177$ & $0.0103$ \\
    U-Net                            & Noise2Void~\cite{krull2019noise2void} & $1.95\mathrm{M}$ & 30.54 \,($+0.29{\pm}0.04$) & $0.883{\pm}0.018$ & $0.0307$ & $0.198$ & $0.0104$ \\
    DnCNN~\cite{zhang2017beyond}     & Noise2Void~\cite{krull2019noise2void} & $0.56\mathrm{M}$ & 30.19 \,($-0.06{\pm}0.03$) & $0.875{\pm}0.018$ & $0.0321$ & $0.224$ & $0.0116$ \\
    CTformer~\cite{wang2022ctformer} & Noise2Void~\cite{krull2019noise2void} & $0.40\mathrm{M}$ & 29.89 \,($-0.36{\pm}0.23$) & $0.862{\pm}0.015$ & $0.0328$ & $0.269$ & $0.0118$ \\
    \bottomrule
  \end{tabular}
\end{table*}

Supervised RED-CNN, U-Net, and DnCNN gain about $4$\,dB over the input and finish
within $0.3$\,dB of one another despite a roughly threefold difference in
parameter count (Table~\ref{tab:results}). CTformer ($+1.77$\,dB) and CFM
($+2.84$\,dB) achieve smaller gains under the same training budget.

Noise2Sim improves PSNR for all four backbones, with gains of
$1.69$--$1.83$\,dB for the CNNs and $0.79$\,dB for CTformer. Noise2Void gives
smaller gains on RED-CNN and U-Net ($0.42$ and $0.29$\,dB) and small losses on
DnCNN and CTformer. This pattern is consistent with the difficulty of applying
pixel-independent blind-spot training to correlated CT noise. The best
label-free result in the main benchmark remains about $2$\,dB below supervised
training. Section~\ref{sec:method} examines whether different pair construction
and flow-based estimation can reduce this gap.

\begin{figure*}[t]
  \centering
  \includegraphics[width=\textwidth]{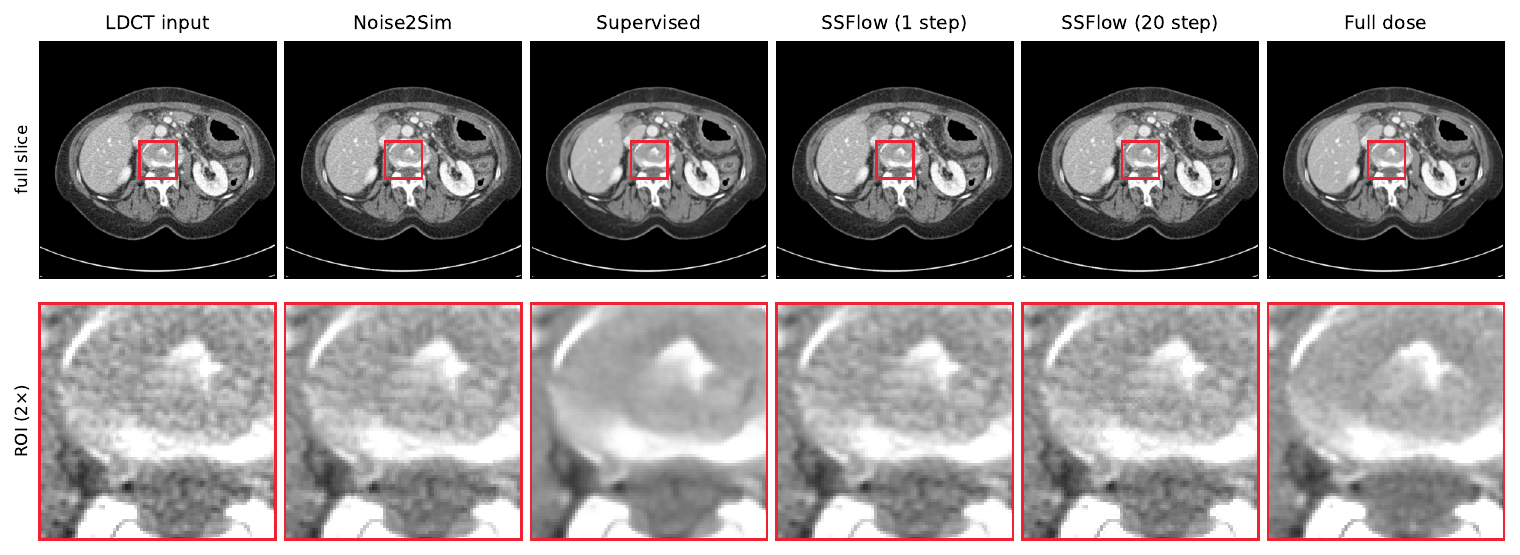}
\caption{Comparison on a held-out abdomen slice. The top row shows the full slice and
ROI location; the bottom row shows a $2\times$ enlargement. Supervised
denoising removes more noise but produces a smoother, lower-frequency texture.
Noise2Sim retains finer texture. Multi-step SSFlow leaves more noise than its
one-step estimate. The NPS-centroid measurements in
Section~\ref{sec:detect} quantify the texture shift.}
  \label{fig:qual}
\end{figure*}

\paragraph{Per-fold stability.}
A 15-epoch replication on the same three folds gives a similar ordering
(Appendix~\ref{app:supp-tables}, Table~\ref{tab:cv}). Supervised CNNs gain
$4.0$--$4.2$\,dB, and Noise2Sim gives the largest label-free improvement
($1.74\pm0.09$\,dB on RED-CNN). The remaining methods have gains or losses of
at most $0.32$\,dB. Supervised RED-CNN reaches $34.43$\,dB at 15 epochs,
compared with $34.50$\,dB at 50 epochs.

\subsection{Task-based detectability}
\label{sec:detect}

\begin{figure*}[t]
  \centering
  \includegraphics[width=\textwidth]{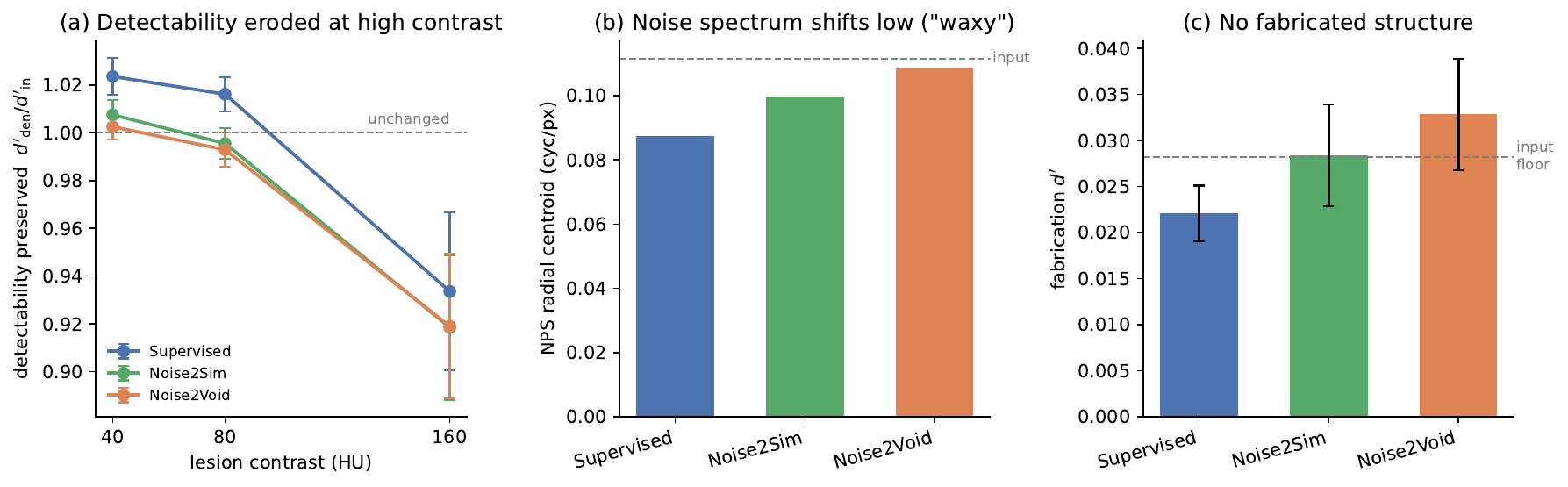}
\caption{Task-based evaluation averaged over three CNN architectures and three seeds.
(a) The preserved-detectability ratio $d'_\text{den}/d'_\text{in}$ falls most
at $160$\,HU, the highest contrast tested. (b) The denoised NPS radial centroid
is lower than the input reference (dashed), particularly for supervised
models. (c) Signal-absent fabrication $d'$ remains close to the input
reference (dashed) across supervision regimes.}
  \label{fig:detect}
\end{figure*}

We evaluate lesion detectability with a channelized Hotelling observer (CHO)
on a signal-known-exactly/background-known-statistically (SKE/BKS) task.
Lesions are inserted in the image domain: a signal-present image is
$x_\text{low}+s$, and the corresponding signal-absent image is
$x_\text{low}$. This preserves the acquired background and its correlated
reconstruction noise without requiring a CT forward projector. The paired
full-dose image provides the reference condition.

We insert Gaussian lesions of radius $4$\,px and contrast $40/80/160$\,HU at
flat in-body sites, using six sites per slice and 200 slices per run. We extract
signal-present and signal-absent ROIs from the input, denoised, and reference
images, apply ten Laguerre-Gauss channels, and compute $d'$. Before use, the
observer is checked against the white-noise expression
$d'=\lVert s\rVert/\sigma$.

We report results for the three CNN architectures in Table~\ref{tab:results},
averaged over architectures and seeds. CTformer shows a similar detectability
loss but is excluded from the CNN aggregate. Per-image methods are omitted
because repeating their optimisation for each observer realisation would be
costly. Signal-absent fabrication measurements for SSFlow are reported
separately in Section~\ref{sec:ss-pairs}.

\begin{table*}[t]
  \centering
  \small
  \caption{CHO evaluation across three CNN architectures, three supervision regimes,
and three seeds (27 runs). Reference $d'$ values are shown above;
preserved-detectability ratios $d'_\text{den}/d'_\text{in}$ below are
mean$\pm$standard deviation within each regime. Ratios above/below one
indicate increased/reduced detectability. We use ratios to account for
between-split variation in input $d'$ (s.d.\ about $1.3$ at $160$\,HU).
The NPS centroid measures noise frequency; a lower value indicates a shift
toward lower frequencies. Fabrication $d'$ is measured on signal-absent
ROIs, with an input reference of $0.028$.}
  \label{tab:detect}
  \begin{tabular}{lccccc}
    \toprule
     & \multicolumn{3}{c}{lesion contrast (HU)} & NPS & fab. \\
    \cmidrule(lr){2-4}
     & $40$ & $80$ & $160$ & centroid\,$\downarrow$ & $d'\,\downarrow$ \\
    \midrule
    \multicolumn{6}{l}{\emph{Reference $d'$ (method-independent)}} \\
    \quad input (floor)   & $1.51$ & $3.03$ & $6.06$ & $0.112$ & $0.028$ \\
    \quad clean (ceiling) & $1.54$ & $3.09$ & $6.18$ & --- & --- \\
    \midrule
    \multicolumn{6}{l}{\emph{Detectability preserved $d'_\text{den}/d'_\text{in}$}} \\
    \quad Supervised & $1.024\,{\pm}\,0.008$ & $1.016\,{\pm}\,0.007$ & $0.934\,{\pm}\,0.033$ & $0.087\,{\pm}\,0.004$ & $0.022\,{\pm}\,0.003$ \\
    \quad Noise2Sim  & $1.008\,{\pm}\,0.006$ & $0.995\,{\pm}\,0.006$ & $0.919\,{\pm}\,0.030$ & $0.100\,{\pm}\,0.005$ & $0.028\,{\pm}\,0.006$ \\
    \quad Noise2Void & $1.003\,{\pm}\,0.006$ & $0.993\,{\pm}\,0.007$ & $0.919\,{\pm}\,0.030$ & $0.109\,{\pm}\,0.006$ & $0.033\,{\pm}\,0.006$ \\
    \bottomrule
  \end{tabular}
\end{table*}

The full-dose reference has only about $2\%$ higher detectability than the noisy
input: at $160$\,HU, $d'=6.18$ versus $6.06$ (Table~\ref{tab:detect}). There
is therefore little room to improve this task at the tested dose. The main
question is how much of the input's detectability survives denoising.

At $160$\,HU, detectability is lower after denoising in all 27 runs. Mean
preserved-detectability ratios are $0.934$ for supervised models and $0.919$
for both Noise2Sim and Noise2Void, corresponding to losses of roughly
$7$--$8\%$. At $40$--$80$\,HU, changes are much smaller, and some ratios
exceed one. In particular, supervised models show mean ratios of $1.024$ and
$1.016$ at these contrasts. The results therefore support a contrast-dependent
effect, rather than a uniform loss of detectability.

Signal-absent fabrication $d'$ remains near the input value
($0.022$--$0.033$ versus $0.028$). This measure does not indicate a clear
increase in fabricated lesion-like signals. Supervised denoising also shifts
the NPS centroid from $0.112$ to $0.087$, a $22\%$ reduction, compared with
$0.109$ for Noise2Void. This shift reflects the smoother, lower-frequency
texture visible in Figure~\ref{fig:qual}.

The variation within each regime is large relative to the differences between
regimes: the $160$\,HU ratios are $0.934\pm0.033$, $0.919\pm0.030$, and
$0.919\pm0.030$. These results do not establish a clear detectability ranking
among the three regimes. Ratios to each run's input reduce the effect of
between-split variation in absolute $d'$ and make the denoiser's effect easier
to compare. Together with the NPS measurements, they show changes that the
aggregate PSNR and SSIM gains do not describe.

% -----------------------------------------------------------------------

\section{One-Step Flow Denoising}
\label{sec:method}

We now compare unconditional rectified flow with direct regression on noisy
pairs. The optimal flow velocity yields the Noise2Noise posterior mean in a
single step. Under jointly Gaussian assumptions, integrating the flow before
applying its denoising readout increases mean squared error. We implement the
one-step estimator as SSFlow and use non-local similarity pairs to reduce the
effect of spatial noise correlation. The ablations compare exclusion radii,
inference step counts, and direct regression on the same pairs.

\subsection{Rectified flow as a Noise2Noise estimator}
\label{sec:rf-n2n}
Let $s$ be the (unobserved) clean slice and let $x_0=s+n_0,\ x_1=s+n_1$ be
two noisy observations with $\mathbb{E}[n_i\mid s]=0$ and
$n_0\perp n_1\mid s$.  Using the same straight-line interpolant
$z_t=(1-t)x_0+t\,x_1$ as Eq.~\eqref{eq:rectified-target}, we train an
\emph{unconditional} velocity network $v_\theta(z,t)$ (inputs: the
interpolant $z$ and time $t$ only, in contrast to the conditional field of
Eq.~\eqref{eq:cfm-loss}) with

\begin{equation}
  \mathcal{L}(\theta)=
  \mathbb{E}_{s,n_0,n_1,\,t\sim\mathcal{U}[0,1]}
  \big\|\,v_\theta(z_t,t)-(x_1-x_0)\,\big\|^2 .
  \label{eq:ss-loss}
\end{equation}

\paragraph{Proposition.}
The minimiser of \eqref{eq:ss-loss} is the conditional expectation
$v^\star(z,t)=\mathbb{E}[\,x_1-x_0\mid z_t=z\,]$.  At $t=0$, $z_0=x_0$, so

\begin{equation}
  x_0 + v^\star(x_0,0)
  = \mathbb{E}[x_1\mid x_0]
  = \mathbb{E}[s\mid x_0],
  \label{eq:ss-mmse}
\end{equation}

the MMSE denoiser. By the tower property,
$\mathbb{E}[x_1\mid x_0]=\mathbb{E}\!\left[\mathbb{E}[x_1\mid s,x_0]\mid
x_0\right]=\mathbb{E}\!\left[s+\mathbb{E}[n_1\mid s,x_0]\mid x_0\right]
=\mathbb{E}[s\mid x_0]$, where $\mathbb{E}[n_1\mid s,x_0]=\mathbb{E}[n_1\mid
s]=0$ because $n_1\perp n_0\mid s$ (so conditioning on $x_0=s+n_0$ adds
nothing beyond $s$) and $\mathbb{E}[n_i\mid s]=0$.  A single velocity
evaluation thus recovers the Noise2Noise~\cite{lehtinen2018noise2noise}
optimum from noisy pairs alone.

This result combines the flow-matching conditional-mean
identity~\cite{gagneux2025phases,albergo2023stochastic} with the Noise2Noise
identity~\cite{lehtinen2018noise2noise}. The target $x_1$ may be noisy as long
as its noise is conditionally zero-mean and independent of the input noise.
The squared-error objective is essential because its minimiser is the
conditional mean. Our goal is restoration by this mean; noisy-data generative
methods such as Ambient Diffusion~\cite{daras2023ambient} instead aim to sample
from a clean-image distribution. Equation~\eqref{eq:ss-mmse} is a
population-level statement. The experiments in Section~\ref{sec:ss-pairs}
measure how closely trained networks approach the corresponding direct
regression result.

\paragraph{Conditioning on the noisy input.}
For $t>0$, giving the network $x_0$ as well as $(z_t,t)$ makes the noisy target
recoverable as $x_1=(z_t-(1-t)x_0)/t$. The network can then fit the pair's
realised noise rather than average over pairs consistent with $z_t$. We omit
$x_0$ as a conditioning channel to avoid this shortcut during training. The
resulting velocity is a conditional average over noisy pairs, as required by
the derivation above.

\subsection{Similarity-based pairs and inference}
\label{sec:ss-pairs}
Equation~\eqref{eq:ss-mmse} requires noise that is independent between the two
observations given the signal. Filtered backprojection (FBP) spreads each
detector reading across neighbouring pixels, so pairs formed from nearby pixels
can retain correlated noise
(Neighbor2Neighbor~\cite{huang2021neighbor2neighbor},
ZS-N2N~\cite{mansour2023zsn2n}). We use the non-local construction of
Noise2Sim~\cite{niu2020noise2sim}: for each query patch, we retrieve its $K$
most similar patches elsewhere in the slice while excluding a radius $r$
around the query. Similarity seeks to match the underlying signal; spatial
separation seeks to reduce noise correlation. These pairs approximate the
assumptions of Eq.~\eqref{eq:ss-mmse}, with $r$ controlling the trade-off
between signal similarity and noise decorrelation.

At inference, the one-step estimate is $\hat s=x+v_\theta(x,0)$. We compare
this estimate with multi-step Euler inference initialised at the noisy input.

\paragraph{Departure from the MMSE estimate in a Gaussian model.}
The one-step identity does not imply that following the learned flow improves
denoising. Its terminal distribution is the noisy law $p_1$. To analyse the
effect of integration, consider a jointly Gaussian model with signal variance
$\sigma_s^2>0$ and independent noise variance $\sigma_n^2>0$ on each
observation, applied per spectral mode. The optimal velocity $v^\star$ is
linear, and the ODE $\dot z=v^\star(z,t)$ has solution
$\hat z(t)=x_0\sqrt{(\sigma_s^2+\gamma(t)\sigma_n^2)/(\sigma_s^2+\sigma_n^2)}$,
where $\gamma(t)=(1-t)^2+t^2$.

The readout $z_t+(1-t)v^\star(z_t,t)=\mathbb{E}[x_1\mid z_t]$ reduces to
the one-step estimate at $t=0$. Evaluating it at the propagated point
$\hat z(t)$ gives the shrinkage estimator $\hat s(t)=\kappa(t)x_0$, with

\begin{equation}
  \kappa(t)=\frac{\sigma_s^2+t\,\sigma_n^2}
                 {\sqrt{(\sigma_s^2+\gamma(t)\sigma_n^2)\,(\sigma_s^2+\sigma_n^2)}}.
  \label{eq:kappa}
\end{equation}

\emph{Proposition (Gaussian flow readout).}  $\kappa$ is strictly increasing on
$[0,1]$, from the Wiener/MMSE coefficient
$\kappa(0)=\sigma_s^2/(\sigma_s^2+\sigma_n^2)$ to $\kappa(1)=1$ (the noisy input); hence the denoising risk
$R(t)=(\kappa(t)-1)^2\sigma_s^2+\kappa(t)^2\sigma_n^2$ is strictly increasing in
$t$.  The readout at $t=0$ is therefore optimal; increasing integration time moves
the readout toward the noisy input.

\emph{Proof.}  Differentiating
$\kappa(t)^2(\sigma_s^2+\sigma_n^2)=(\sigma_s^2+t\sigma_n^2)^2/(\sigma_s^2+\gamma(t)\sigma_n^2)$,
the numerator of the derivative is
$2\sigma_n^2(\sigma_s^2+t\sigma_n^2)(2\sigma_s^2+\sigma_n^2)(1-t)\ge0$, vanishing
only at $t{=}1$, so $\kappa$ strictly increases; $R$ is convex in $\kappa$ with
minimiser $\kappa(0)$, hence strictly increasing in $t$.\hfill$\square$

For every $t>0$, this readout applies less shrinkage than the MMSE solution.
The result concerns the exact Gaussian flow and its readout; the behaviour of
a learned field with a finite Euler step size is an empirical question. It
also suggests why better velocity fitting need not improve multi-step
denoising: more accurate transport still targets a noisy endpoint.

\paragraph{Ablation.}
We vary the exclusion radius $r\in\{1,2,3\}$ and the number of inference
steps $\{1,4,8,20\}$ (Table~\ref{tab:ssflow-steps}). We also compare
one-step flow with direct regression while holding the pairs fixed
(Table~\ref{tab:flow-vs-reg}).

One-step SSFlow performs best at $r=2$, with a $1.88$\,dB PSNR gain. Gains
are $0.13$\,dB lower at $r=1$, which excludes only the query pixel and
recovers the plain Noise2Sim target, and $0.19$\,dB lower at $r=3$. This
suggests that excluding nearby pixels helps, but that a larger exclusion
radius can also reduce pair quality.

\begin{figure*}[t]
  \centering
  \includegraphics[width=\textwidth]{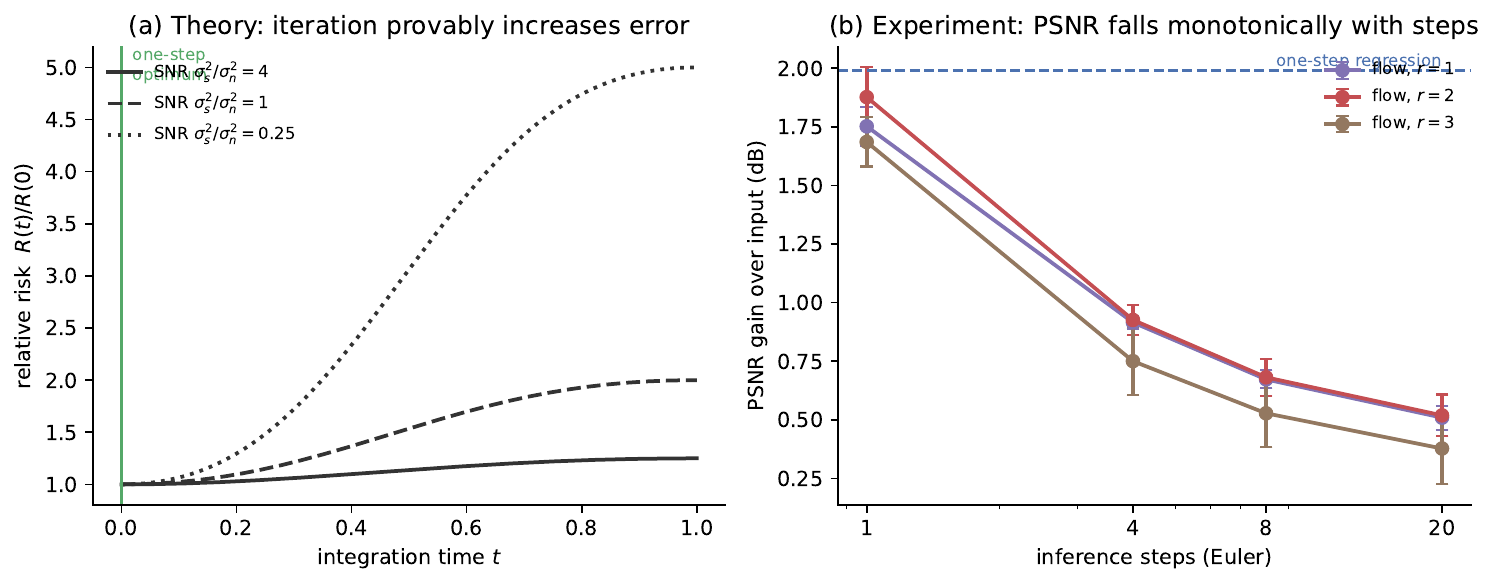}
  \\[4pt]
  \includegraphics[width=\textwidth]{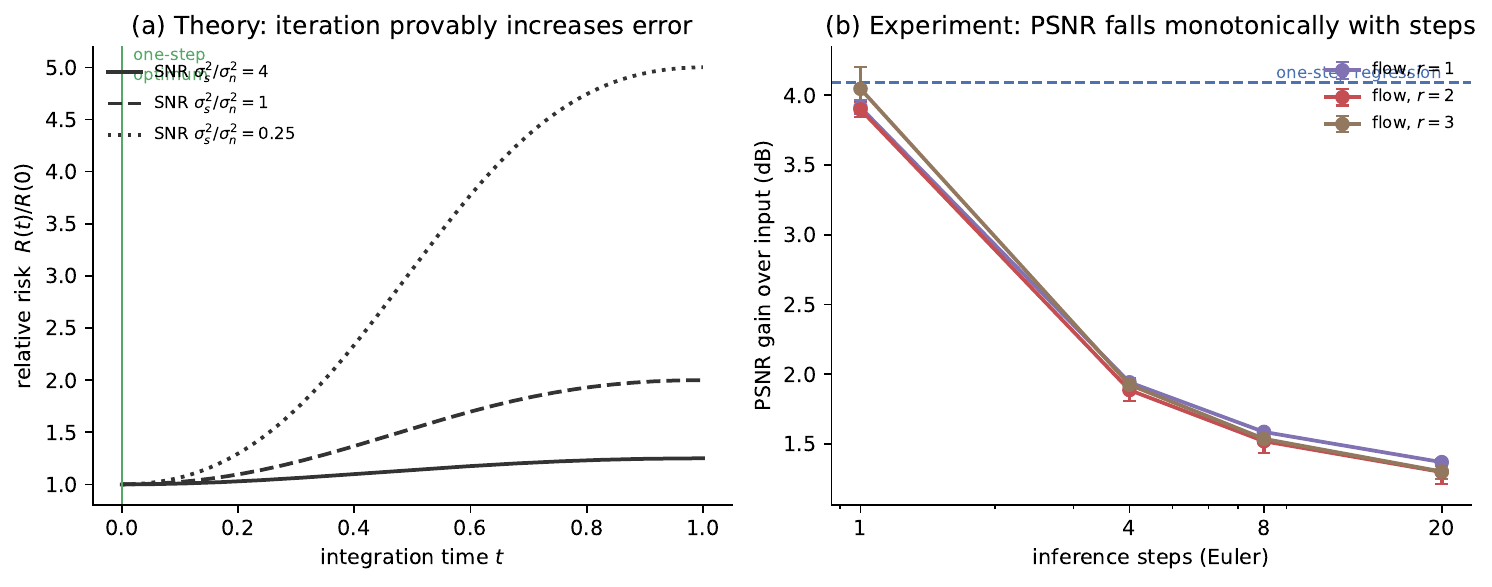}
  \caption{Integration and denoising performance. (a) In the jointly Gaussian model,
relative risk $R(t)/R(0)$ increases with integration time
(Eq.~\eqref{eq:kappa}). (b) On CT, PSNR gains decrease as the Euler step
count increases. The dashed line marks the matched-pair regression result.
(c) BSDS500 images with independent Gaussian noise show the same trend.
Empirical results use three seeds; configurations are
\texttt{sweep\_hallucination.yml} and
\texttt{sweep\_natural\_hallucination.yml}.}
  \label{fig:steps}
\end{figure*}

Additional inference steps reduce PSNR (Figure~\ref{fig:steps}). At $r=2$,
the gain falls from $1.88$\,dB with one step to $0.93$, $0.68$, and
$0.52$\,dB with 4, 8, and 20 steps. The other radii show the same direction
of change. Validation PSNR also peaks early in training and then falls while
GMSD continues to improve; we select the checkpoint with the highest
validation PSNR.

On identical pairs, one-step flow and direct regression differ by at most
$0.2$\,dB. Regression performs slightly better at each radius, reaching
$1.99$\,dB at $r=2$ compared with $1.88$\,dB for flow. Both estimators
perform best at the same radius. Thus the pair construction benefits both
methods, and these experiments provide no fidelity advantage for flow
matching. Direct regression at $r=2$ gives our best label-free CT result.
The supervised comparison also favours MSE regression: U-Net exceeds CFM by
about $1$\,dB (Section~\ref{sec:results}).

These comparisons concern the reported fidelity metrics and are consistent
with matched-data CT benchmarks~\cite{eulig2024benchmark}. They do not
establish a ranking on perceptual or radiomic-feature measures, which may
reward different aspects of image texture.

\paragraph{Natural-image experiment.}
The one-step identity depends on conditional noise independence and the
squared-error objective, rather than CT-specific assumptions. We test the
same SSFlow model on BSDS500 natural images corrupted with white Gaussian
noise ($\sigma=0.1$), using the same exclusion-radius grid and three seeds
(Table~\ref{tab:natural}). Direct regression exceeds one-step flow by
$0.15$--$0.21$\,dB. Increasing the Euler step count again reduces PSNR,
with gains falling from about $3.9$ to $1.3$\,dB over 1 to 20 steps.
The exclusion radius has a smaller effect than on CT, consistent with the
absence of spatial noise correlation. These results extend the empirical
one-step advantage to an independent-noise setting.

\begin{table}[t]
  \centering
  \small
  \caption{BSDS500 natural images with white Gaussian noise ($\sigma=0.1$).
Values are PSNR gains over the noisy input, averaged over three seeds.
Direct regression slightly exceeds one-step flow at each exclusion radius,
while additional Euler steps reduce the flow's gain.}
  \label{tab:natural}
  \begin{tabular}{ccccccc}
    \toprule
    & \multicolumn{2}{c}{one-step $\Delta$PSNR} & \multicolumn{4}{c}{flow, Euler steps} \\
    \cmidrule(lr){2-3}\cmidrule(lr){4-7}
    $r$ & flow & regression & $1$ & $4$ & $8$ & $20$ \\
    \midrule
    $1$ & $+3.91$ & $+4.12$ & $+3.91$ & $+1.94$ & $+1.59$ & $+1.37$ \\
    $2$ & $+3.90$ & $+4.09$ & $+3.90$ & $+1.89$ & $+1.52$ & $+1.30$ \\
    $3$ & $+4.05$ & $+4.20$ & $+4.05$ & $+1.92$ & $+1.54$ & $+1.30$ \\
    \bottomrule
  \end{tabular}
\end{table}

\subsection{Ablation results}
\label{sec:sweep-tables}
Table~\ref{tab:ssflow-steps} reports the inference-step and exclusion-radius
ablation (\texttt{sweep\_hallucination.yml}). Table~\ref{tab:flow-vs-reg}
compares flow and regression on the same pairs
(\texttt{sweep\_flow\_vs\_reg.yml}). Per-image results appear in
Appendix~\ref{app:supp-tables}, Table~\ref{tab:sweep-periimage}.

\begin{table*}[t]
  \centering
  \small
  \caption{SSFlow on TCIA LDCT abdomen data across inference step counts and exclusion
radii (three seeds, 50 epochs; \texttt{sweep\_hallucination.yml}).
PSNR includes the gain over the mean $30.25$\,dB input. Additional steps
reduce PSNR; signal-absent fabrication $d'$ remains near the input value
of $0.028$. \textbf{Bold}: best PSNR, SSIM, RMSE, and RSR.
$\uparrow$/$\downarrow$: higher/lower is better.}
  \label{tab:ssflow-steps}
  \begin{tabular}{ccrccccc}
    \toprule
    \textbf{Steps} & \textbf{Radius $r$} &
    PSNR\,$\uparrow$ (\,$\Delta$) & SSIM\,$\uparrow$ & RMSE\,$\downarrow$ &
    GMSD\,$\downarrow$ & RSR\,$\downarrow$ & fab.\ $d'$\,$\downarrow$ \\
    \midrule
    \multicolumn{8}{l}{\emph{Reference}} \\
    \quad LDCT input & --- & 30.25 & 0.876 & 0.0320 & 0.174 & 0.0115 & 0.028 \\
    \midrule
    \multicolumn{8}{l}{\emph{One-step posterior mean}} \\
    $1$ & $1$ & 32.00 \,(+1.75) & 0.899 & 0.0261 & 0.215 & 0.0077 & 0.030 \\
    $1$ & $2$ & \textbf{32.12 \,(+1.88)} & \textbf{0.902} & \textbf{0.0256} & 0.213 & \textbf{0.0073} & 0.029 \\
    $1$ & $3$ & 31.93 \,(+1.69) & 0.896 & 0.0261 & 0.240 & 0.0076 & 0.028 \\
    \midrule
    \multicolumn{8}{l}{\emph{Four-step Euler}} \\
    $4$ & $1$ & 31.17 \,(+0.92) & 0.889 & 0.0288 & 0.206 & 0.0095 & 0.030 \\
    $4$ & $2$ & 31.18 \,(+0.93) & 0.890 & 0.0287 & 0.204 & 0.0093 & 0.028 \\
    $4$ & $3$ & 31.00 \,(+0.75) & 0.887 & 0.0292 & 0.229 & 0.0096 & 0.028 \\
    \midrule
    \multicolumn{8}{l}{\emph{Eight-step / Twenty-step Euler}} \\
    $8$ & $2$ & 30.93 \,(+0.68) & 0.887 & 0.0295 & 0.205 & 0.0099 & 0.028 \\
    $20$ & $2$ & 30.77 \,(+0.52) & 0.885 & 0.0301 & 0.205 & 0.0103 & 0.028 \\
    \bottomrule
  \end{tabular}
\end{table*}

\begin{table*}[t]
  \centering
  \small
  \caption{Direct regression and one-step SSFlow trained on identical similarity pairs.
The regressor is RED-CNN ($1.85\mathrm{M}$ parameters); the flow velocity
network has $2.11\mathrm{M}$ parameters. Each estimator uses three radii and
three seeds (nine runs). PSNR gains are measured against each split's input;
uncertainties are standard deviations across seeds. Regression has a small
advantage at each radius, and both methods perform best at $r=2$.
Signal-absent fabrication $d'$ remains near the input value. Configurations:
\texttt{sweep\_flow\_vs\_reg.yml} and \texttt{sweep\_hallucination.yml}.
\textbf{Bold}: best PSNR. $\uparrow$/$\downarrow$: higher/lower is better.}
  \label{tab:flow-vs-reg}
  \begin{tabular}{lcrccccc}
    \toprule
    \textbf{Estimator} & \textbf{Radius $r$} &
    PSNR\,$\uparrow$ (\,$\Delta$) & SSIM\,$\uparrow$ & RMSE\,$\downarrow$ &
    GMSD\,$\downarrow$ & RSR\,$\downarrow$ & fab.\ $d'$\,$\downarrow$ \\
    \midrule
    \multicolumn{8}{l}{\emph{Reference}} \\
    \quad LDCT input & --- & 30.25 & 0.876 & 0.0320 & 0.174 & 0.0115 & 0.028 \\
    \midrule
    \multicolumn{8}{l}{\emph{Regression (Noise2Sim, RED-CNN) --- same pairs, direct MMSE}} \\
    Regression & $1$ & 32.02 \,($+1.77{\pm}0.14$) & $0.900{\pm}0.015$ & $0.0260$ & $0.163$ & $0.0076$ & $0.029$ \\
    Regression & $2$ & \textbf{32.24 \,($+1.99{\pm}0.09$)} & $0.904{\pm}0.015$ & $0.0253$ & $0.160$ & $0.0071$ & $0.030$ \\
    Regression & $3$ & 32.12 \,($+1.87{\pm}0.11$) & $0.904{\pm}0.014$ & $0.0256$ & $0.162$ & $0.0072$ & $0.029$ \\
    \midrule
    \multicolumn{8}{l}{\emph{Flow (SSFlow, one-step) --- same pairs, rectified-flow objective}} \\
    Flow & $1$ & 32.00 \,($+1.75{\pm}0.08$) & $0.899{\pm}0.013$ & $0.0261$ & $0.215$ & $0.0077$ & $0.030$ \\
    Flow & $2$ & 32.12 \,($+1.88{\pm}0.13$) & $0.902{\pm}0.016$ & $0.0256$ & $0.213$ & $0.0073$ & $0.029$ \\
    Flow & $3$ & 31.93 \,($+1.69{\pm}0.10$) & $0.896{\pm}0.011$ & $0.0261$ & $0.240$ & $0.0076$ & $0.028$ \\
    \bottomrule
  \end{tabular}
\end{table*}

% -----------------------------------------------------------------------
\section{Discussion}

\paragraph{Architecture and objective.}
Under the common training budget, the three supervised CNNs finish within
$0.3$\,dB of one another despite a roughly threefold parameter range.
CTformer and CFM achieve smaller gains. U-Net and CFM have similar network
bodies, yet MSE regression exceeds flow matching by about $1$\,dB. The
label-free comparison likewise gives no PSNR advantage to one-step flow over
regression on the same pairs. These results favour direct regression for
fidelity under the tested settings; they do not establish that the same
ordering holds at other training budgets.

\paragraph{Metric selection.}
PSNR and SSIM measure agreement with a reference, but can disagree with
perceptual assessments of medical
images~\cite{breger2025reassess,lee2025ldctiqa}. GMSD provides a
gradient-based comparison~\cite{xue2014gradient}, and RSR measures relative
residual energy. Task-based observers~\cite{li2021detection,kc2024lgcho} and
hallucination measures~\cite{bhadra2021halluc,tivnan2024halluc} address
different questions. Here, the CHO detects a loss at $160$\,HU despite
improvements in aggregate PSNR and SSIM. Reporting observer performance and
noise texture alongside fidelity metrics gives a fuller account of the
denoiser's effect.

\paragraph{Limitations.}
The three-seed CT benchmark and detectability study use the abdomen window.
The supervision ordering also appears with simulated low-dose noise
(Table~\ref{tab:sim-ldct}) and on chest images (Table~\ref{tab:chest}), but
the chest experiment uses only one seed. The per-image methods are reported
on a single split in Table~\ref{tab:sweep-periimage}, with a separate
reduced-budget comparison in Table~\ref{tab:cv}; they are omitted from the
observer study because of repeated optimisation costs.

The one-step identity assumes conditionally independent, zero-mean noise.
Non-local similarity pairs approximate these assumptions and can still differ
in underlying signal. The increasing-risk result is proved for the exact
jointly Gaussian flow and its readout. A general bound for learned fields
and finite Euler steps remains open, although the CT and natural-image
experiments show declining PSNR with additional steps.

The observer study uses synthetic lesions inserted into reconstructed images
and is not a reader study. Its signal-absent measurements also address only
the tested lesion-like signals; they do not rule out other forms of fabricated
structure. A broader evaluation of generative CT denoisers would require
additional detection tasks and hallucination measures.

% -----------------------------------------------------------------------

\section{Conclusion}

An unconditional rectified flow trained on Noise2Noise pairs yields the
posterior-mean denoiser in one step at the population optimum. In a jointly
Gaussian model, integrating the flow before applying its readout increases
denoising risk. CT and natural-image experiments show a similar pattern:
one-step flow performs close to direct regression, and additional Euler
steps reduce PSNR. Direct regression on spatially separated, similar patches
gives the best label-free CT result, gaining $1.99$\,dB over the input.
The CTDenoiser benchmark also shows why fidelity metrics alone are
insufficient: supervised models gain about $4$\,dB, yet the observer detects
a loss of lesion detectability at the highest tested contrast. Code and
experiment configurations will be released publicly.

\bibliographystyle{plain}

% ===== Supplementary material: after references; not counted in the 7-page limit =====
\appendix

\section{Robustness experiments}
\label{app:robustness}
We repeat the benchmark with a simulated low-dose noise source and a second
anatomy to assess how the results depend on the data.

\subsection{Robustness to a second LDCT source}
\label{sec:sim-ldct}
The main benchmark uses the TCIA low/full-dose pairs. For a second noise
source, we simulate low-dose images from the same full-dose abdomen volumes
with a correlated, signal-dependent noise model
(\texttt{-{}-sim-base-std=0.03}, \texttt{-{}-sim-signal-std=0.06}).
This gives an input PSNR of about $28.3$\,dB, compared with $30.25$\,dB for
the original data. Patient splits, patch size, optimiser, seeds, and epochs
are held fixed (\texttt{sweep\_sim\_ldct.yml}).

The supervision ordering persists (Table~\ref{tab:sim-ldct}). Supervised
CNNs gain $4.5$--$4.9$\,dB and finish within $0.4$\,dB of one another.
Noise2Sim improves all four backbones, with gains near $0.9$\,dB for the
CNNs. Noise2Void remains close to the noisy baseline for the CNNs
($-0.09$ to $+0.04$\,dB) and below it for CTformer. The smaller Noise2Sim
gains show that the size of the label-free improvement depends on the noise
source, even when the method ordering is unchanged.

\begin{table*}[t]
  \centering
  \small
  \caption{Benchmark with simulated, correlated, signal-dependent noise added to the
full-dose abdomen volumes (\texttt{sweep\_sim\_ldct.yml}; 50 epochs,
matched patient splits). PSNR includes gains over the $28.26$\,dB input.
Results use three seeds except the marked CTformer entries:
$^{\dagger}$supervised training has two completed seeds, and Noise2Void
has one, following failures near epochs 47--49. Single-seed entries have
no standard deviation. \textbf{Bold}: best $\Delta$PSNR within each
supervision regime. $\uparrow$/$\downarrow$: higher/lower is better.}
  \label{tab:sim-ldct}
  \begin{tabular}{llrccccc}
    \toprule
    \textbf{Model} & \textbf{Supervision} & \textbf{Params} &
    PSNR\,$\uparrow$ (\,$\Delta$) & SSIM\,$\uparrow$ & RMSE\,$\downarrow$ &
    GMSD\,$\downarrow$ & RSR\,$\downarrow$ \\
    \midrule
    Simulated LDCT input & --- (no denoising) & --- & $28.26$ & $0.597$ & $0.0388$ & $0.418$ & $0.0153$ \\
    \midrule
    RED-CNN~\cite{chen2017lowdose}   & Supervised & $1.85\mathrm{M}$ & \textbf{33.17 \,($+4.91{\pm}0.13$)} & $0.937$ & $0.0221$ & $0.169$ & $0.0050$ \\
    DnCNN~\cite{zhang2017beyond}     & Supervised & $0.56\mathrm{M}$ & 32.90 \,($+4.64{\pm}0.14$) & $0.934$ & $0.0228$ & $0.222$ & $0.0053$ \\
    U-Net                            & Supervised & $1.95\mathrm{M}$ & 32.78 \,($+4.52{\pm}0.11$) & $0.933$ & $0.0231$ & $0.222$ & $0.0055$ \\
    CTformer$^{\dagger}$~\cite{wang2022ctformer} & Supervised & $0.40\mathrm{M}$ & 30.22 \,($+1.96{\pm}0.04$) & $0.860$ & $0.0309$ & $0.361$ & $0.0098$ \\
    CFM~\cite{lipman2022flow}        & Supervised & $2.11\mathrm{M}$ & 31.07 \,($+2.81{\pm}0.17$) & $0.914$ & $0.0281$ & $0.226$ & $0.0081$ \\
    \midrule
    RED-CNN~\cite{chen2017lowdose}   & Noise2Sim~\cite{niu2020noise2sim} & $1.85\mathrm{M}$ & \textbf{29.22 \,($+0.96{\pm}0.07$)} & $0.706$ & $0.0347$ & $0.409$ & $0.0123$ \\
    U-Net                            & Noise2Sim~\cite{niu2020noise2sim} & $1.95\mathrm{M}$ & 29.19 \,($+0.93{\pm}0.05$) & $0.673$ & $0.0348$ & $0.384$ & $0.0124$ \\
    DnCNN~\cite{zhang2017beyond}     & Noise2Sim~\cite{niu2020noise2sim} & $0.56\mathrm{M}$ & 29.11 \,($+0.85{\pm}0.05$) & $0.667$ & $0.0351$ & $0.395$ & $0.0126$ \\
    CTformer~\cite{wang2022ctformer} & Noise2Sim~\cite{niu2020noise2sim} & $0.40\mathrm{M}$ & 28.81 \,($+0.55{\pm}0.09$) & $0.810$ & $0.0364$ & $0.346$ & $0.0135$ \\
    \midrule
    RED-CNN~\cite{chen2017lowdose}   & Noise2Void~\cite{krull2019noise2void} & $1.85\mathrm{M}$ & \textbf{28.30 \,($+0.04{\pm}0.09$)} & $0.643$ & $0.0386$ & $0.398$ & $0.0152$ \\
    DnCNN~\cite{zhang2017beyond}     & Noise2Void~\cite{krull2019noise2void} & $0.56\mathrm{M}$ & 28.24 \,($-0.02{\pm}0.04$) & $0.627$ & $0.0389$ & $0.390$ & $0.0154$ \\
    U-Net                            & Noise2Void~\cite{krull2019noise2void} & $1.95\mathrm{M}$ & 28.17 \,($-0.09{\pm}0.08$) & $0.646$ & $0.0391$ & $0.381$ & $0.0157$ \\
    CTformer$^{\dagger}$~\cite{wang2022ctformer} & Noise2Void~\cite{krull2019noise2void} & $0.40\mathrm{M}$ & 28.04 \,($-0.20$) & $0.723$ & $0.0397$ & $0.342$ & $0.0163$ \\
    \bottomrule
  \end{tabular}
\end{table*}

\subsection{Robustness to a second anatomy}
\label{sec:chest}
We also evaluate real low/full-dose chest pairs from ten patients, with eight
assigned to training and two to validation. This experiment uses one seed
(Table~\ref{tab:chest}). Supervised training gives the largest PSNR gain for
each CNN. RED-CNN gains $3.29$\,dB over the $39.68$\,dB input under
supervision and $1.52$\,dB with Noise2Sim. Noise2Void gives losses for all
three CNNs.

Supervised RED-CNN preserves $0.46$ of the input's low-contrast detectability
despite its PSNR improvement. These results require caution because the input
PSNR is higher than in the abdomen study and only one split is tested.
CTformer diverged on this training set, with all modes below the input
baseline, and is omitted from the table. Replication across additional
splits is needed to assess variability.

\begin{table*}[t]
  \centering
  \small
  \caption{Chest LDCT benchmark (\texttt{sweep\_chest.yml}): real low/full-dose pairs
from ten patients, with eight training and two validation patients; one seed.
PSNR includes gains over the $39.68$\,dB input. CTformer diverged and is
omitted; CFM is evaluated only under supervision. \textbf{Bold}: best
$\Delta$PSNR within each supervision regime.
$\uparrow$/$\downarrow$: higher/lower is better.}
  \label{tab:chest}
  \begin{tabular}{llrccccc}
    \toprule
    \textbf{Model} & \textbf{Supervision} & \textbf{Params} &
    PSNR\,$\uparrow$ (\,$\Delta$) & SSIM\,$\uparrow$ & RMSE\,$\downarrow$ &
    GMSD\,$\downarrow$ & RSR\,$\downarrow$ \\
    \midrule
    Chest LDCT input & --- (no denoising) & --- & $39.68$ & $0.932$ & $0.0110$ & $0.235$ & $0.0004$ \\
    \midrule
    RED-CNN~\cite{chen2017lowdose}   & Supervised & $1.85\mathrm{M}$ & \textbf{42.97 \,($+3.29$)} & $0.970$ & $0.0075$ & $0.209$ & $0.0002$ \\
    CFM~\cite{lipman2022flow}        & Supervised & $2.11\mathrm{M}$ & 41.97 \,($+2.29$) & $0.965$ & $0.0084$ & $0.206$ & $0.0002$ \\
    DnCNN~\cite{zhang2017beyond}     & Supervised & $0.56\mathrm{M}$ & 41.70 \,($+2.02$) & $0.959$ & $0.0088$ & $0.216$ & $0.0002$ \\
    U-Net                            & Supervised & $1.95\mathrm{M}$ & 41.01 \,($+1.33$) & $0.965$ & $0.0092$ & $0.207$ & $0.0002$ \\
    \midrule
    RED-CNN~\cite{chen2017lowdose}   & Noise2Sim~\cite{niu2020noise2sim} & $1.85\mathrm{M}$ & \textbf{41.20 \,($+1.52$)} & $0.954$ & $0.0092$ & $0.219$ & $0.0003$ \\
    U-Net                            & Noise2Sim~\cite{niu2020noise2sim} & $1.95\mathrm{M}$ & 40.09 \,($+0.41$) & $0.959$ & $0.0102$ & $0.212$ & $0.0003$ \\
    DnCNN~\cite{zhang2017beyond}     & Noise2Sim~\cite{niu2020noise2sim} & $0.56\mathrm{M}$ & 39.85 \,($+0.17$) & $0.935$ & $0.0107$ & $0.232$ & $0.0004$ \\
    \midrule
    U-Net                            & Noise2Void~\cite{krull2019noise2void} & $1.95\mathrm{M}$ & \textbf{39.62 \,($-0.05$)} & $0.959$ & $0.0107$ & $0.212$ & $0.0003$ \\
    RED-CNN~\cite{chen2017lowdose}   & Noise2Void~\cite{krull2019noise2void} & $1.85\mathrm{M}$ & 39.38 \,($-0.30$) & $0.949$ & $0.0111$ & $0.223$ & $0.0003$ \\
    DnCNN~\cite{zhang2017beyond}     & Noise2Void~\cite{krull2019noise2void} & $0.56\mathrm{M}$ & 38.03 \,($-1.65$) & $0.929$ & $0.0129$ & $0.240$ & $0.0005$ \\
    \bottomrule
  \end{tabular}
\end{table*}

\section{Supplementary Tables}
\label{app:supp-tables}

Table~\ref{tab:cv} reports the 15-epoch cross-validation comparison.
Table~\ref{tab:sweep-periimage} gives the single-split per-image results.

\begin{table*}[t]
  \centering
  \small
  \caption{Three-fold patient-level cross-validation on TCIA LDCT abdomen data
(15 epochs). $\Delta$PSNR and SSIM are mean$\pm$standard deviation;
PSNR, RMSE, and RSR are fold means. Input PSNR is
$31.44/30.04/29.27$\,dB for folds 1/2/3. DnCNN and RED-CNN have complete
fold coverage; U-Net is omitted because its coverage is incomplete.
Filter2Noise and ZS-N2N are per-image methods.
\textbf{Bold}: best per column. \underline{Underline}: best label-free
PSNR gain.}
  \label{tab:cv}
  \begin{tabular}{llccccc}
    \toprule
    \textbf{Model} & \textbf{Supervision} &
    $\Delta$PSNR\,$\uparrow$ (dB) & PSNR\,$\uparrow$ & SSIM\,$\uparrow$ &
    RMSE\,$\downarrow$ & RSR\,$\downarrow$ \\
    \midrule
    RED-CNN~\cite{chen2017lowdose} & Supervised & \textbf{+4.19 $\pm$ 0.22} & \textbf{34.43} & \textbf{0.927 $\pm$ 0.013} & \textbf{0.0195} & \textbf{0.0042} \\
    DnCNN~\cite{zhang2017beyond}   & Supervised & +4.04 $\pm$ 0.22 & 34.29 & 0.924 $\pm$ 0.015 & 0.0199 & 0.0044 \\
    \midrule
    RED-CNN~\cite{chen2017lowdose} & Noise2Sim~\cite{niu2020noise2sim} & \underline{+1.74 $\pm$ 0.09} & 31.99 & 0.900 $\pm$ 0.015 & 0.0261 & 0.0077 \\
    DnCNN~\cite{zhang2017beyond}   & Noise2Sim~\cite{niu2020noise2sim} & +1.42 $\pm$ 0.28 & 31.66 & 0.895 $\pm$ 0.015 & 0.0271 & 0.0082 \\
    \midrule
    RED-CNN~\cite{chen2017lowdose} & Noise2Void~\cite{krull2019noise2void} & +0.32 $\pm$ 0.10 & 30.57 & 0.883 $\pm$ 0.019 & 0.0306 & 0.0105 \\
    DnCNN~\cite{zhang2017beyond}   & Noise2Void~\cite{krull2019noise2void} & $-0.02$ $\pm$ 0.06 & 30.23 & 0.875 $\pm$ 0.018 & 0.0320 & 0.0114 \\
    \midrule
    Filter2Noise~\cite{sun2025filter2noise} & Zero-shot & +0.26 $\pm$ 0.11 & 30.51 & 0.881 $\pm$ 0.016 & 0.0309 & 0.0105 \\
    ZS-N2N~\cite{mansour2023zsn2n}          & Zero-shot & $-0.09$ $\pm$ 0.20 & 30.16 & 0.879 $\pm$ 0.015 & 0.0319 & 0.0112 \\
    \bottomrule
  \end{tabular}
\end{table*}

\begin{table*}[t]
  \centering
  \small
  \caption{Per-image methods evaluated on split 0 (\texttt{sweep\_periimage.yml}).
PSNR includes gains over that split's $31.44$\,dB input. Both methods
remain close to the input baseline.
$\uparrow$/$\downarrow$: higher/lower is better.}
  \label{tab:sweep-periimage}
  \begin{tabular}{llrccccc}
    \toprule
    \textbf{Method} & \textbf{Type} & \textbf{Params} &
    PSNR\,$\uparrow$ (\,$\Delta$) & SSIM\,$\uparrow$ & RMSE\,$\downarrow$ &
    GMSD\,$\downarrow$ & RSR\,$\downarrow$ \\
    \midrule
    Filter2Noise~\cite{sun2025filter2noise} & Zero-shot & --- & 31.58 \,(+0.14) & 0.896 & 0.0269 & 0.176 & 0.0076 \\
    ZS-N2N~\cite{mansour2023zsn2n}          & Zero-shot & $21.3\mathrm{K}$ & 31.17 \,($-0.27$) & 0.892 & 0.0281 & 0.182 & 0.0083 \\
    \bottomrule
  \end{tabular}
\end{table*}

\section{Reproducibility}
\label{app:repro}
Sections~\ref{sec:rf-n2n} and~\ref{sec:ss-pairs} state the assumptions and
derivations for the one-step identity and Gaussian flow readout. The experiments
use the public TCIA Low-Dose CT and Projection dataset and BSDS500. Data
preprocessing, architectures, training settings, and inference are described in
the Methods section; seed counts and exceptions are specified with each table.

The training loop, model definitions, evaluation code, and experiment
configurations will be released publicly, including the full-slice tiling,
CHO, and NPS implementations. The repository link is withheld for anonymous
review.

\end{document}